\documentclass[letterpaper, 10 pt, conference]{ieeeconf}
\IEEEoverridecommandlockouts

\newif\ifarxiv
\arxivtrue

\def\MYTITLE{Real-world Latency Analysis\\of Vehicular Visible Light Communication \\with Multiple LED Transmitters and an Event-Based Camera}

\ifarxiv
\usepackage[breaklinks,colorlinks]{hyperref}
\hypersetup{
  pdftitle={\MYTITLE},
  pdfsubject={Computer Vision, Robotics},
  pdfauthor={Ryota Soga, Tsukasa Shimizu, Shan Lu, Shintaro Shiba, Quan Kong, Shan Lu, Takaya Yamazato},
  pdfkeywords={Event Cameras, Visible Light Communication, Optical Communication}
}
\usepackage[absolute]{textpos}
\else
\usepackage[pagebackref,breaklinks,colorlinks]{hyperref}
\fi

\usepackage[T1]{fontenc} 
\usepackage{cite}
\usepackage{amsmath,amssymb,amsfonts}
\usepackage{algorithmic}
\usepackage{graphicx}
\usepackage{textcomp}
\usepackage{xcolor}
\usepackage{ragged2e}
\usepackage{url}
\usepackage{setspace}
\usepackage{amsmath}
\usepackage{orcidlink}
\usepackage[capitalize]{cleveref}
\crefname{section}{Sec.}{Secs.}
\Crefname{section}{Section}{Sections}
\Crefname{figure}{Figure}{Figures}
\crefname{figure}{Fig.}{Figs.}
\Crefname{table}{Table}{Tables}
\crefname{table}{Tab.}{Tabs.}

\def\BibTeX{{\rm B\kern-.05em{\sc i\kern-.025em b}\kern-.08em
    T\kern-.1667em\lower.7ex\hbox{E}\kern-.125emX}}
    
\begin{document}

\ifarxiv
\definecolor{somegray}{gray}{0.5}
\newcommand{\darkgrayed}[1]{\textcolor{somegray}{#1}}
\begin{textblock}{11.5}(2.25, 0.3)  
\begin{center}
\darkgrayed{This paper has been accepted for publication at the IEEE VTC2026-Spring, \\
Nice, France, 2026.
\copyright IEEE}
\end{center}
\end{textblock}
\fi

\title{\MYTITLE}

\author{%
\centering
\parbox{0.9\textwidth}{\centering
Ryota Soga$^{1}$, Tsukasa Shimizu$^{2}$, Shintaro Shiba$^{3}$, Quan Kong$^{3}$, Shan Lu$^{1}$, Takaya Yamazato$^{1}$\\[2mm]
\small
$^{1}$Graduate School of Engineering, Nagoya University, Nagoya, Japan\\
\{soga, lu, yamazato\}@yamazato.nuee.nagoya-u.ac.jp\\
$^{2}$TOYOTA MOTOR CORPORATION, Toyota, Japan\\
tsukasa\_shimizu@mail.toyota.co.jp\\
$^{3}$Woven by Toyota, Inc., Tokyo, Japan\\
\{shintaro.shiba, quan.kong\}@woven.toyota
}%
}


\maketitle

\begin{abstract}
Event cameras offer high temporal resolution, low latency, and wide dynamic range, making them promising receivers for visible light communication (VLC) in vehicle-to-everything (V2X) applications. This work presents an event-camera-based VLC system addressing three key challenges: bandwidth saturation, multi-transmitter reception, and latency characterization. 

We adopt a positive-event-only mode and design a protocol that suppresses event generation while maintaining communication distance and a wide field of view. We also propose a method to identify multiple transmitters and demonstrate simultaneous reception from up to three LEDs. Finally, we evaluate end-to-end latency in real vehicular scenarios and show that the system meets cooperative perception requirements. These results demonstrate that event-camera-based VLC is a feasible complement to existing V2X technologies (e.g., RF).
\end{abstract}


\section{Introduction}
\label{sec:introduction}


Event-based vision sensors, or event cameras, are emerging image sensors that provide high temporal resolution (in the order of microseconds), efficient data rates, and a wide dynamic range (over 120~dB)~\cite{Gallego2022pami}.
Each pixel independently monitors brightness and outputs an ``event'' consisting of a timestamp, polarity (positive for dark-to-light transitions, negative for light-to-dark transitions), and pixel coordinates as soon as an intensity change exceeds a predefined threshold.
Their high spatio-temporal resolution makes event cameras attractive for visible light communication (VLC) receivers, compared with traditional ones such as photodiodes and frame-based (CCD/CMOS) sensors.
Also, they are suitable for vehicle-to-everything (V2X) communication,
since ($i$) they naturally capture fast LED blinking as the binary signals,
($ii$) they have the potential to achieve communication in extremely low end-to-end latency,
and ($iii$) their spatial resolution is suitable for further applications such as localization.
Notably, when combined with RF communication, VLC serves as a complementary technology that alleviates congestion in the RF bands.

Several works have explored optical communications using event cameras for moving targets:
Shen et al.~\cite{shen1,shen2} demonstrate 16~kbps communication using vehicle headlights, and von Arnim et al.~\cite{anrim} evaluate a drone-mounted transmitter with tracking. 
However, research on event-camera-based VLC in vehicular environments remains limited, especially in scenarios where latency is critical.
To fully leverage these advantages, it is paramount to develop a system that evaluates end-to-end latency as a real-world performance metric~\cite {Maqueda2018Steering}.

\begin{figure}[t!]
    \centering
    \includegraphics[width = 0.8\linewidth]{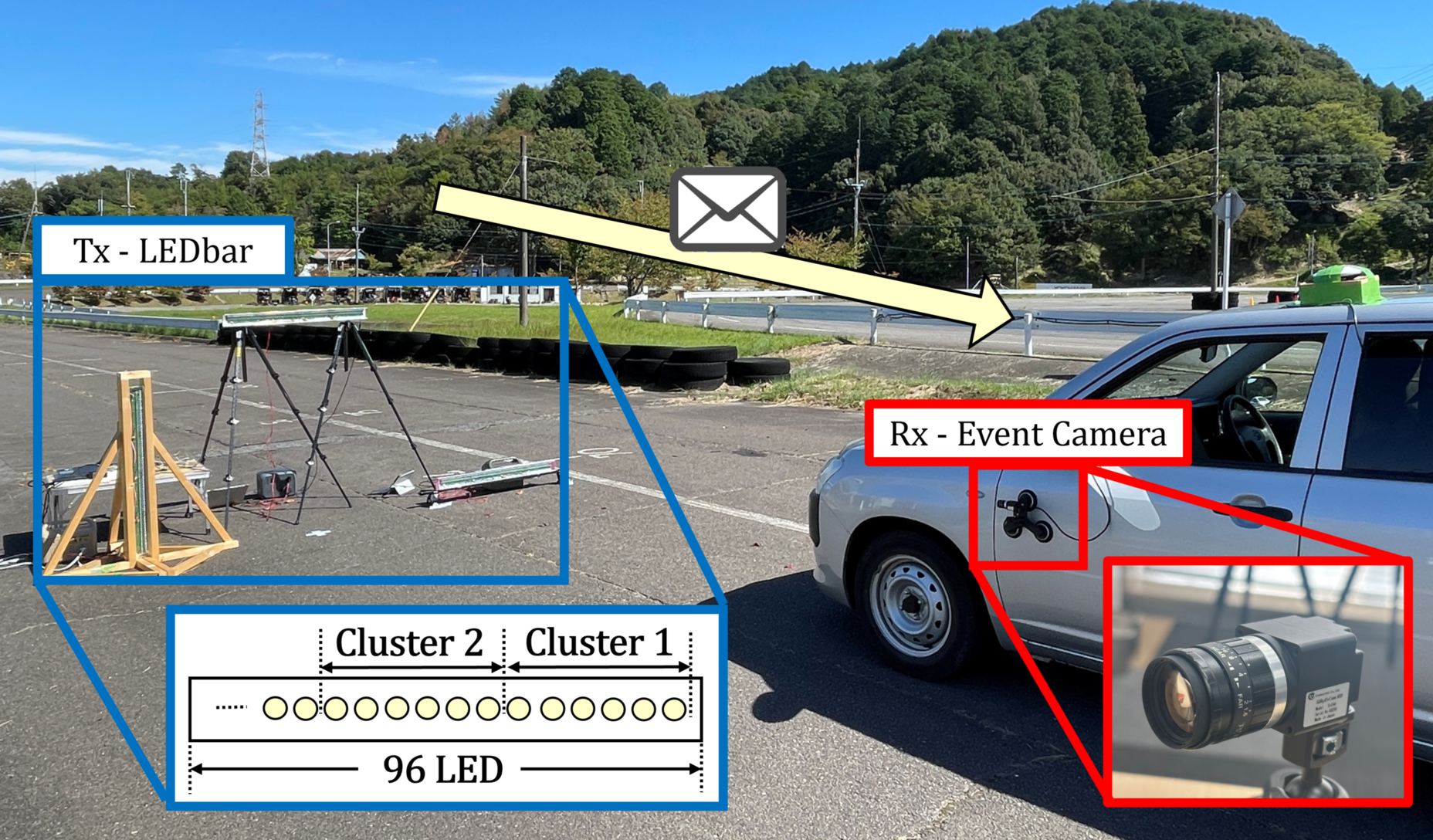}
    \vspace{-2mm}
    \caption{We evaluated the end-to-end latency of infrastructure-to-vehicle VLC using an event camera. Experiments in a realistic multi-transmitter vehicular environment confirm that the system meets cooperative perception requirements and demonstrate its effectiveness.}
    \vspace{-5mm}
    \label{fig:rec-and-tra}
\end{figure}

In our previous work, we investigated robust protocols and the communication capabilities of event-camera VLC in detail~\cite{soga2026}. 
We also demonstrate high-throughput VLC with densely arranged LEDs by mitigating multiple-LED interference and validate the system in controlled vehicular environments.
Nevertheless, three challenges remain towards the real-world deployment.

The first challenge is the degradation of communication performance caused by bandwidth saturation in the event-camera data. 
Vehicle-mounted cameras can generate numerous events at high speed, and the event rate may exceed the data-transfer bandwidth (e.g., USB cables), leading to event loss and communication failures. Prior approaches increase the lens aperture or set the region of interest (ROI) in the image plane~\cite{apcc}, but they reduce the communication distance or narrow the field of view. In particular, narrowing the field of view makes it difficult to capture multiple transmitters distributed across the image frame. To tackle this trade-off, we design a communication protocol optimized for positive-event-only operation, inspired by Shiba et al.~\cite{E-VLC-CVPRW-2025},
which propose interval-based protocols based solely on positive events.
It suppresses event generation while maintaining communication distance and preserving a wide field of view.

Second, conventional methods are limited to receiving signals from only a single transmitter. 
In real-world vehicular environments, however, multiple LEDs are typically present in the field of view, making multi-transmitter reception indispensable. 
In this work, we present a transmitter identification method and experimentally demonstrate simultaneous reception from up to three transmitters.
The third challenge is the lack of end-to-end latency evaluation.
Latency analysis is essential for ITS applications, such as cooperative perception. 
However, few studies analyze latency in VLC systems, and none exist for optical camera communication (OCC) systems. 
Nakano et al.~\cite{nakano} evaluate the latency using a high-speed camera; however, their evaluation is limited to static conditions and a single transmitter. In contrast, this study conducts experimental latency measurements in an on-vehicle environment with multiple transmitters, enabled by addressing the two challenges described above, and evaluates the effectiveness of the proposed system for cooperative perception, a V2X use case.

Our results demonstrate that the proposed system receives 288 bytes within 100 ms, exceeding the latency requirements set by the European Telecommunications Standards Institute (ETSI), which specifies approximately 200 bytes transmitted within 100 ms for cooperative perception. These results demonstrate the effectiveness of our system and present the first benchmark for latency evaluation in event-camera-based VLC, an area that has not been sufficiently established to date.

The contributions of this paper are summarized as follows:
\begin{itemize}
    \item We design a communication protocol optimized for a positive-event-only operational mode, which suppresses event-camera bandwidth overflow and enables extended communication distance while maintaining a wide field of view.
    \item We develop a method for identifying multiple transmitters within the field of view and enable simultaneous reception and evaluation of signals from up to three transmitters.
    \item We measure end-to-end communication latency and demonstrate that the proposed system satisfies the requirements for cooperative perception in vehicular environments.
\end{itemize}

\vspace{-1mm}
\section{System Model}
\label{sec:system_model}

\subsection{Transmitter and Receiver}
\vspace{-1mm}
Fig.~\ref{fig:rec-and-tra} shows the transmitter and receiver.  
The transmitter employs a vertical LED bar consisting of 96 red LEDs, each operating at a blinking frequency of 10 kHz. These LEDs are grouped into clusters of six, referred to as LED Clusters, and each LED Cluster transmits a different waveform. On–off keying modulation is used for waveform transmission.

For the receiver, we use the SilkyEvCam equipped with the SONY IMX636 sensor (temporal resolution below 100~\textmu s, resolution 1280~$\times$~720). Using the Metavision SDK, the threshold for generating negative events is set to 190, the maximum value, to prevent them. Furthermore, the voltage high-pass filter in the pixel circuit was set to its maximum value of 120 to cut low-frequency components, while all other bias parameters were kept at their default value of 0. With these settings, the system operates under conditions that significantly reduce the number of generated events. In practice, under this configuration, no data output exceeding the camera’s data-transfer bandwidth was observed.

\vspace{-1mm}
\subsection{Overview of System model}
\begin{figure}[t!]
\centering
\includegraphics[width=0.8\linewidth]{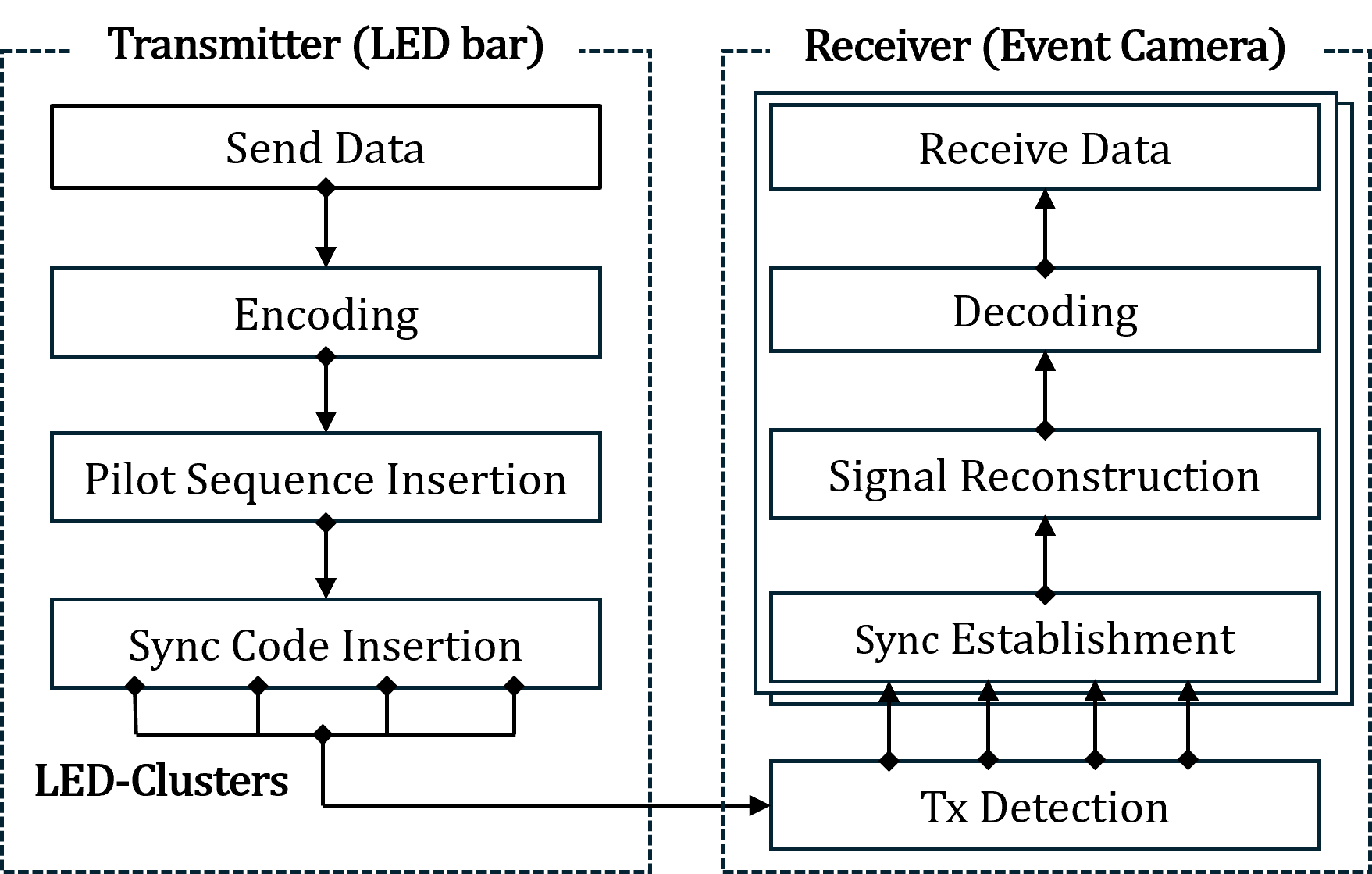}
\vspace{-2mm}
\caption{System model. The system model used in our previous work has been extended to recognize multiple transmitters and to process each transmitter independently. In addition, the communication protocol has been redesigned to handle event-rate saturation.}
\label{fig:systemmodel}
\vspace*{-4mm}
\end{figure}

We describe the overview of the system model used in this study, as shown in Fig.~\ref{fig:systemmodel}.  
On the transmitter side, the transmission data are first encoded, after which a synchronization sequence is appended at the beginning, and pilot sequences are inserted between segments of the transmission data.  

On the receiver side, the system first roughly identifies the region where each transmitter is located. Events are then separated for each detected region, and subsequent processing is performed individually for the events corresponding to each transmitter. In the actual implementation, a dedicated thread is assigned to each transmitter so that the subsequent processing can be executed in parallel.

In the following processing stage, synchronization is established in the same manner as in our previous work~\cite{soga}, after which the pilot sequences are used to reconstruct the waveform of each LED Cluster. For this reconstruction, the correlation between the received events at each pixel and the pilot sequence (indicating how accurately the pilot is received at that pixel) is used as a weight when combining the pixel-wise waveforms. Finally, decoding is performed.

The next sections describe in detail the components of the system that have been modified specifically for this study.

\vspace{-2mm}

\subsection{Communication Protocol and Decoding Algorithm}
In our previous work, we assigned a value of ``1'' to positive events and ``$-1$'' to negative events. However, when only positive events are used, this approach becomes infeasible. Moreover, because event cameras generate events only when brightness changes, no events are generated when a sequence of consecutive ``1'' bits is transmitted; therefore, no positive events corresponding to the second and subsequent ``1'' bits are generated. Thus, in our new method, it is necessary to avoid consecutive symbols with the same polarity and to ensure that the waveform can be reconstructed solely from positive events.

To address this issue, we apply Manchester coding and propose a new waveform reconstruction algorithm. In addition, owing to sensor noise and interference from adjacent LEDs, achieving error-free communication is difficult. Therefore, we introduce CRC (Cyclic Redundancy Check)-Polar codes decoded by Successive Cancellation List (SCL) decoding~\cite{wang2020adf_scl}. Furthermore, to fully exploit the error-correction capability, we reconstruct likelihood-inclusive waveforms and perform soft-decision decoding.

\begin{figure}[t!]
    \centering
    \includegraphics[width = 0.95\linewidth]{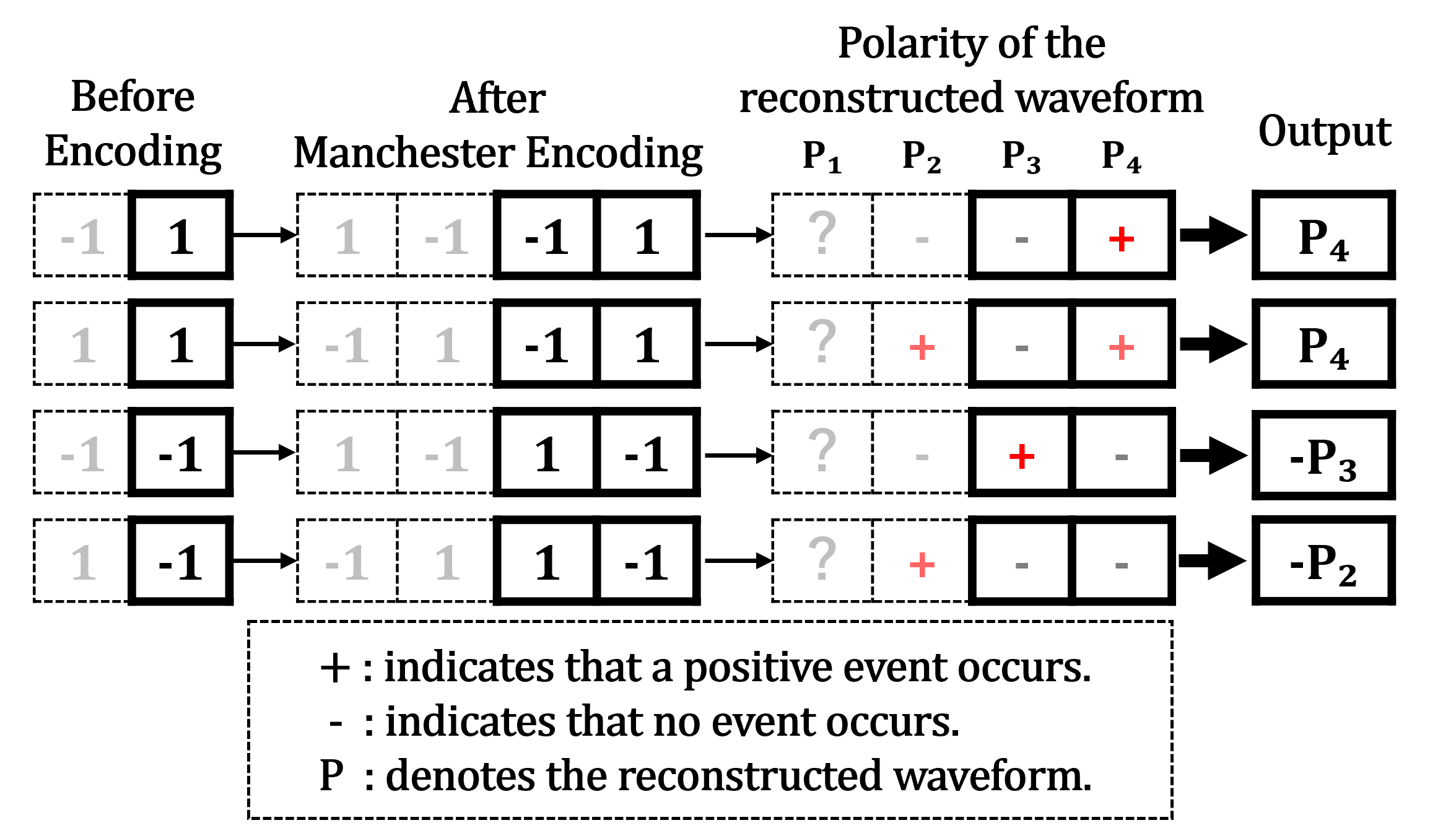}
    \vspace{-3mm}
    \caption{Manchester decoding algorithm. We operate the event camera in a mode that outputs only positive events to avoid event-rate saturation. Under this constraint, we propose a new communication protocol that maintains communication performance without degradation.}
    \vspace{-4mm}
    \label{fig:man-recon}
\end{figure}

An overview of the method is shown in Fig.~\ref{fig:man-recon}. Since event generation depends on the previously transmitted bit, four cases must be considered, as illustrated in the figure. $P1$–$P4$ represent the reconstructed waveforms for each LED Cluster and include likelihood values derived from the number of generated events and the correlation with the pilot sequence. Based on the positions of the positive events produced by Manchester coding, the waveform is reconstructed as illustrated in Fig.~\ref{fig:man-recon}.

Furthermore, due to sensor noise and interference from neighboring LEDs, events may be generated in both $P3$ and $P4$, making it impossible to determine which of the two represents the correct reconstruction. To resolve this ambiguity, we select the waveform with the larger likelihood between $P3$ and $P4$. The reconstructed waveform is then fed into the Polar decoder using soft-decision metrics.

Through this approach, decoding becomes feasible using only positive events. A performance comparison between this method and our previous method is presented in Section \ref{sec:experiment}.

\subsection{Transmitter-Region Identification and Synchronization}
We present examples of a transmitter-region identification algorithm and a synchronization algorithm that operate using only positive events. Both algorithms use the synchronization sequence prepended to each packet. As the synchronization sequence, we adopt a pattern in which the event-generation intervals transition as $5T \rightarrow 4T \rightarrow 3T$, where $T$ denotes the LED blinking period. On the receiver side, the spatial and temporal distributions of events generated by this synchronization sequence are constructed.

\begin{figure}[t!]
\centering
\includegraphics[width = 0.70\linewidth]{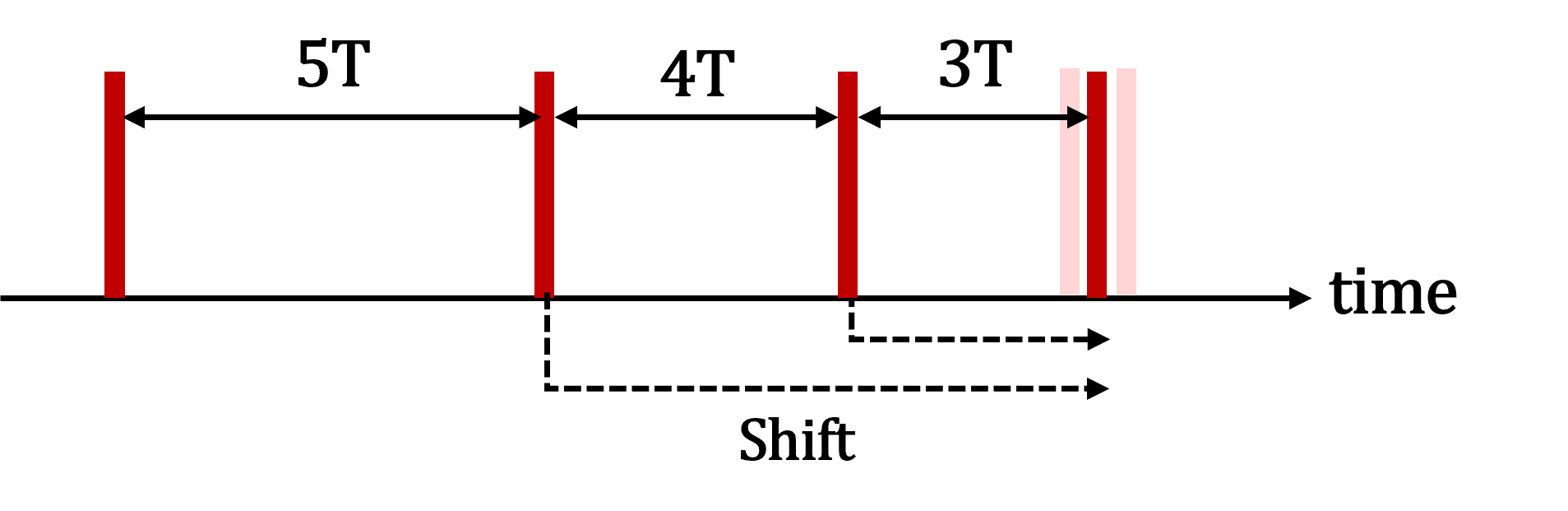}
\vspace{-5mm}
\caption{Example of the synchronization and multi-transmitter identification algorithms operating with positive events only. Common processing steps are shown. Events occurring at specific intervals are time-shifted and recorded as illustrated.}
\label{fig:shift}
\end{figure}

For each output event from the event camera, the receiver stores the timestamp of the most recent event at each pixel. The time interval between the stored timestamp and the newly received event is computed, and if the interval matches $3T$, $4T$, or $5T$, the timestamp is updated. When the interval is $5T$, $4T + 3T$ is added before recording; when the interval is $4T$, $3T$ is added. This procedure aligns events to a single time instance, as illustrated in Fig.~\ref{fig:shift}. This processing is applied to all events within a packet.

\begin{figure}[t!]
\centering
\includegraphics[width = 0.65\linewidth]{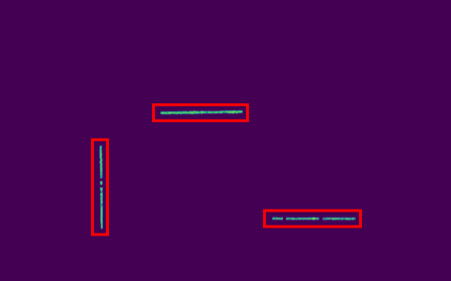}
\vspace{-2mm}
\caption{Example of transmitter-region detection obtained using the multi-transmitter recognition algorithm. Each detected transmitter is enclosed by a rectangular region. In subsequent processing, events within each rectangle are extracted and handled independently in parallel threads.}
\label{fig:detect}
\vspace*{-4mm}
\end{figure}

In the transmitter-region identification algorithm (Fig.~\ref{fig:detect}), the spatial distribution of recorded events is constructed. Gaussian filtering is first applied to suppress noise. Next, the flood-fill method \cite {pfaltz1966} is used to detect clusters of pixels containing recorded events, and bounding rectangles are drawn around each cluster. Rectangles that are spatially close to each other are then merged. Rectangles are subsequently removed based on their area and the number of pixels within them whose recorded-event counts exceed a threshold. The remaining rectangles are judged to contain transmitters. After identification, events within each transmitter region are extracted, and subsequent processing is performed independently for each region. This method identifies transmitter regions by analyzing event distributions that are highly likely to originate from transmitters.

For the synchronization algorithm, the temporal distribution of recorded events is constructed using the timestamping procedure based on the synchronization sequence described at the beginning of this section. By computing the temporal density of this distribution, a clear peak emerges at the aligned time instance because events concentrate at a single time point. The time corresponding to the largest peak is defined as the synchronization timing. This method is compatible with operation modes that use only positive events and enables a shorter synchronization sequence than that used in our previous work.


\vspace{-2mm}
\section{Field Experiments}
\label{sec:experiment}
\vspace{-2mm}
We conducted experimental evaluations of the aforementioned system.  
Data was captured by a vehicle-mounted event camera as it drove past stationary roadside transmitters. The vehicle maintained a straight path, passing at a lateral distance of about 2 meters from the transmitters.

\subsection{BER Characteristics of the Proposed Protocol}
\begin{figure}[t!]
    \centering
    \includegraphics[width = 0.9\linewidth]{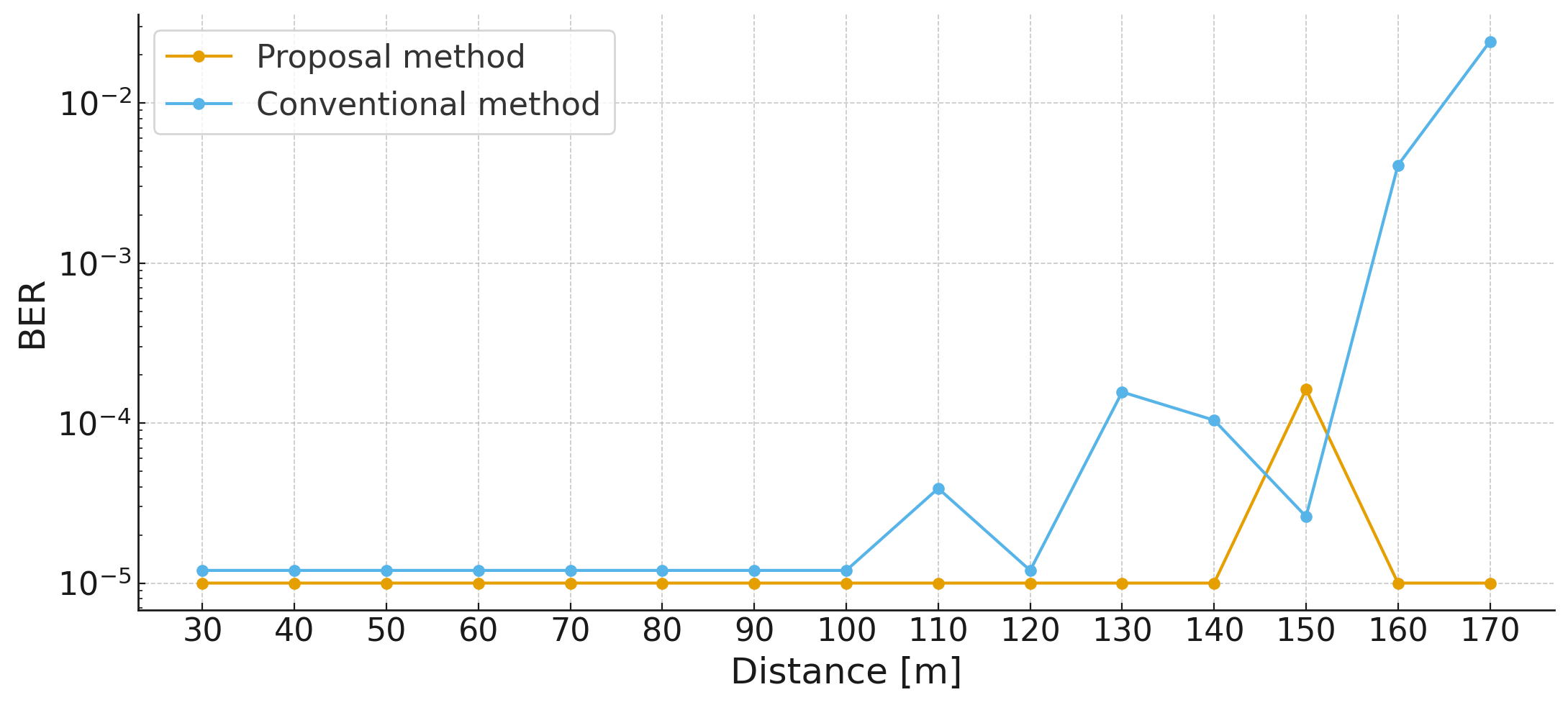}
    \vspace{-3mm}
    \caption{The BER characteristics of the conventional and proposed protocols in a static environment are shown. The results demonstrate that the proposed protocol achieves performance comparable to or better than the conventional protocol under static conditions.}
    \label{fig:stop}
\end{figure}

\begin{figure}[t!]
    \centering
    \includegraphics[width = 0.9\linewidth]{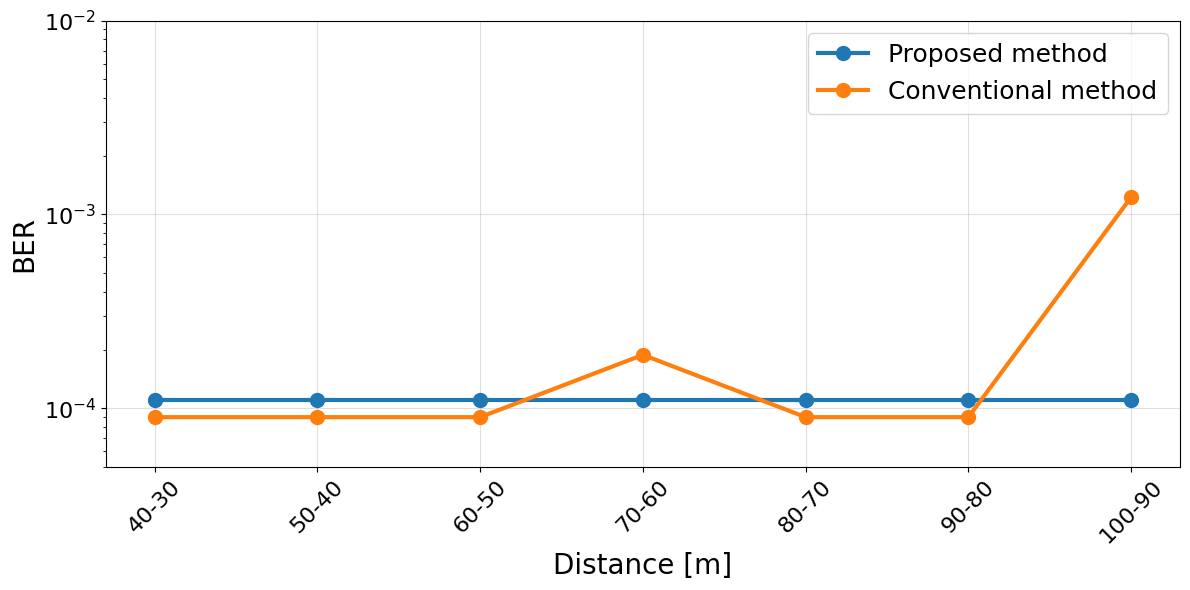}
    \vspace{-3mm}
    \caption{The BER characteristics of the conventional protocol~\cite{apcc} and the proposed protocol during vehicle motion at 40~km/h are shown. While the conventional protocol restricts the ROI during capture, the proposed protocol operates with the entire image enabled and still achieves comparable or superior performance.}
    \label{fig:comparison}
\end{figure}

The BER performance of the proposed protocol was evaluated under both static conditions and vehicle motion at 40 km/h (Fig.~\ref{fig:comparison}, Fig.~\ref{fig:stop}). Static experiments were conducted in the evening under clear skies, and mobile experiments in the daytime under similar conditions. A single transmitter was used in all experiments. In the mobile scenario, BER was averaged over the data received while the vehicle traveled 10 m. The conventional protocol baseline is taken from our previous work~\cite{apcc}. Both protocols use a coding rate of 3/8, while the conventional system restricts the ROI to half of the image to reduce event generation and avoid bandwidth saturation.

Results show that the proposed protocol achieves lower BER at equal or longer distances than the conventional protocol, while enabling full-frame reception. This demonstrates that positive-event-only operation with the proposed protocol maintains performance without ROI limitation.

In addition, during transmitter-region identification, we observed instances where regions outside the transmitter were incorrectly detected. Reception processing was performed even for these falsely identified regions.

\subsection{Measurement of Communication Latency}
We performed real-time processing using the proposed algorithms and measured communication latency under clear daytime conditions. Time synchronization was achieved using GPS with a Pulse Per Second (PPS) signal, resulting in only a few milliseconds of offset.

Latency was defined as the time from the start of transmitter blinking to completion of packet reception at the receiver. It includes command transmission, packet blinking duration, data transfer from the camera to the PC, and reception processing, while encoding and pattern transfer (~1 ms) were excluded as negligible.

\begin{figure}[t!]
    \centering
    \includegraphics[width = 1\linewidth]{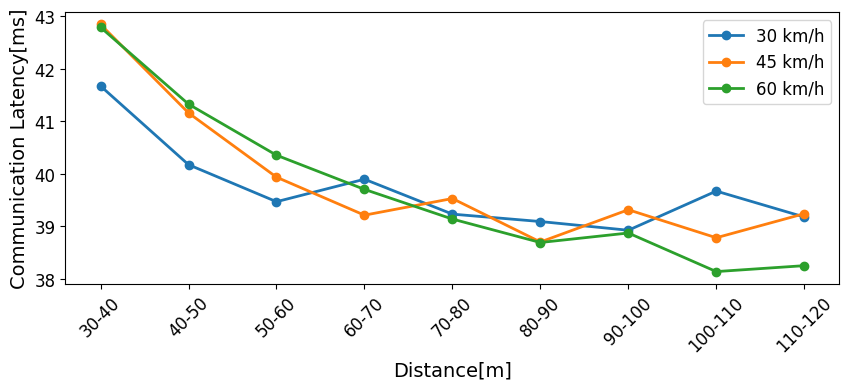}
    \vspace{-3mm}
    \caption{Latency measured while varying vehicle speed (payload: 96 bytes, 1 transmitter). Vehicle speed has no observable impact on latency.}
    \vspace{-4mm}
    \label{fig:speed}
\end{figure}

Fig.~\ref{fig:speed} shows that vehicle speed has little effect on latency. Although higher speeds increase event rates, most events originate from the background and do not significantly impact processing around the transmitter. In contrast, shorter distances increase latency due to a larger apparent transmitter size and higher event density.

\begin{figure}[t!]
    \centering
    \includegraphics[width = 1\linewidth]{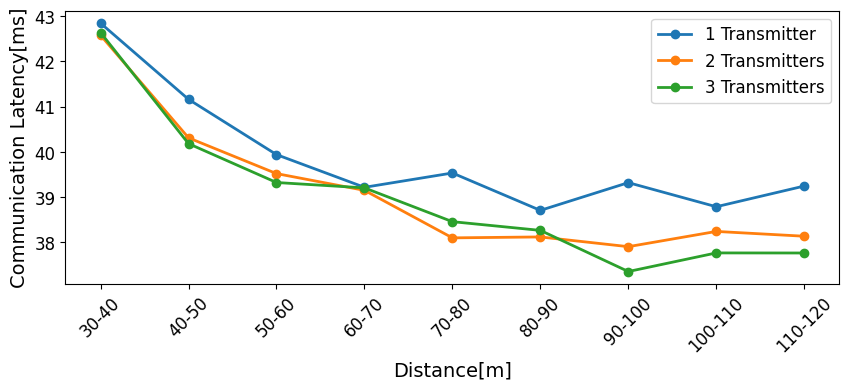}
    \vspace{-3mm}
    \caption{Latency measured while varying the number of transmitters (payload: 96 bytes, speed: 45 km/h). The impact of the number of transmitters on latency is negligible.}
    \vspace{-4mm}
    \label{fig:tx}
\end{figure}

Fig.~\ref{fig:tx} shows that the number of transmitters has minimal impact on latency due to parallel processing. Processing remains stable as long as the number of transmitters is within the CPU core limit, although excessive transmitters may cause event overload and communication failure.

\begin{figure}[t!]
    \centering
    \includegraphics[width = 1\linewidth]{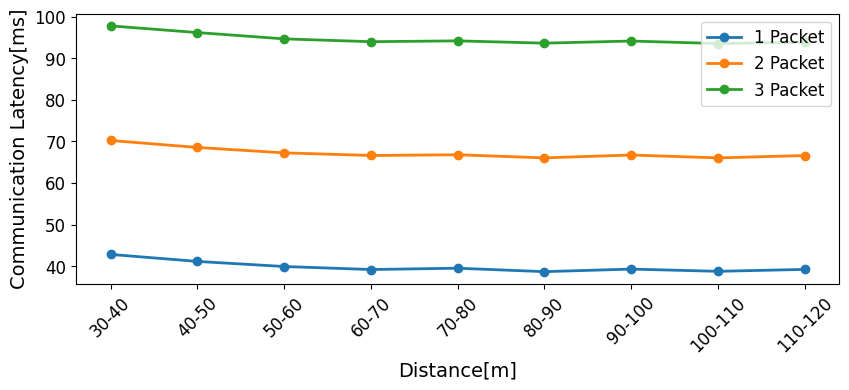}
    \vspace{-3mm}
    \caption{Latency measured while varying payload size (1 transmitter, speed: 45 km/h). Latency increases with payload size, remaining under ~100 ms for 288 bytes.}
    \vspace{-4mm}
    \label{fig:packet}
\end{figure}

Fig.~\ref{fig:packet} shows that latency increases with payload size due to longer blinking duration (27.4 ms per packet), while reception processing time remains unchanged.

Most latency is dominated by LED blinking time, with only 10–16 ms attributed to processing, indicating sufficient computational margin. The packet error rate (PER) during these measurements was not zero, but remained within an acceptable range for latency evaluation.

Figures~\ref{fig:speed}--\ref{fig:packet} show the results of the communication latency measurements.  

\section{Evaluation of Communication Latency}
\label{sec:cp}
One of the key use cases of vehicular communication is cooperative perception, in which vehicles, pedestrians, and infrastructure (e.g., roadside units) share sensor information to detect objects and blind spots that a single agent cannot perceive. 
In this study, we evaluate our system based on the ETSI communication requirements for cooperative perception.

According to the Collective Perception Service defined in ETSI TR~103~562 and ETSI TS~103~324, the CPM (Collective Perception Message) generation interval is specified as 100--1000~ms (nominally 100~ms)~\cite{ETSI562}, and the time from data generation to transmission must not exceed 100~ms for safety-critical applications~\cite{ETSI324}. 
Thus, we set the target communication latency to below 100~ms. 
Since a CPM is typically 200~bytes or larger, a VLC system must complete the reception of at least 200~bytes within this time budget.

In relation to this target, the latency measurements in Fig.~\ref{fig:packet} show that receiving three packets (about 288~bytes) satisfies the 100~ms requirement. 
These results indicate that our VLC system can support cooperative perception.

However, the system requires a line-of-sight (LOS) link between the transmitter and receiver and provides lower data capacity than RF communication. 
Therefore, we consider VLC most effective when used in combination with RF communication as a complementary technology rather than a standalone replacement.

\section{Conclusion}
\label{sec:conclusion}
This paper presented an event-camera-based VLC system designed for practical vehicular environments.  
We addressed bandwidth saturation by adopting a positive-event-only mode and developing a protocol tailored to this operation, thereby mitigating bandwidth saturation while preserving the communication range and ensuring a sufficient field of view to capture multiple transmitters.
We further enabled simultaneous reception from multiple transmitters and evaluated end-to-end latency under realistic driving scenarios.  
To the best of our knowledge, this is the first latency analysis of an event-camera-based OCC system in vehicular settings.  
Experimental results showed that the proposed system extends communication distance while maintaining a wide field of view and meets ETSI latency requirements for cooperative perception.  
Future work will focus on improving transmitter identification robustness and exploring relay-based communication to overcome line-of-sight limitations.


\section*{Acknowledgment}
The authors gratefully acknowledge the partial financial travel support provided by the Hara Research Foundation through a travel grant for international conference presentation.

\vspace{-2mm}

\bibliographystyle{IEEEtran}
\bibliography{soga.bib}

\end{document}